\documentclass[sigconf,natbib=true]{acmart}

\AtBeginDocument{%
  \providecommand\BibTeX{{%
    \normalfont B\kern-0.5em{\scshape i\kern-0.25em b}\kern-0.8em\TeX}}}

\usepackage[normalem]{ulem}
\usepackage{enumitem}

\copyrightyear{2021} 
\acmYear{2021} 
\setcopyright{acmcopyright}
\acmConference[CIKM '21]{Proceedings of the 30th ACM International Conference on Information and Knowledge Management}{November 1--5, 2021}{Virtual Event, QLD, Australia}
\acmBooktitle{Proceedings of the 30th ACM International Conference on Information and Knowledge Management (CIKM '21), November 1--5, 2021, Virtual Event, QLD, Australia}
\acmPrice{15.00}
\acmDOI{10.1145/3459637.3482078}
\acmISBN{978-1-4503-8446-9/21/11}

\begin{document}
\fancyhead{}
 
\title{CrossAug: A Contrastive Data Augmentation Method for Debiasing Fact Verification Models}

\author{Minwoo Lee\textsuperscript{1}, Seungpil Won$^{1}$, Juae Kim$^{2}$, Hwanhee Lee$^{1}$, Cheoneum Park$^{2}$, Kyomin Jung$^{1}$}\authornotemark[1]
\affiliation{%
\institution{$^{1}$Dept. of Electrical and Computer Engineering, Seoul National University, $^{2}$AIRS Company, Hyundai Motor Group
} 
\country{}
}
\email{
{minwoolee, seungpil.won, wanted1007, kjung}@snu.ac.kr, {juaekim, cheoneum.park}@hyundai.com
}

\newcommand{\tony}[1]{#1}
\newcommand{\minwoo}[1]{#1}
\newcommand{\hwanhee}[1]{#1}

\begin{abstract}
Fact verification datasets are typically constructed using crowdsourcing techniques due to the lack of text sources with veracity labels. 
However, the crowdsourcing process often produces undesired biases in data that cause models to learn spurious patterns. 
In this paper, we propose CrossAug\footnote{Code available at https://github.com/minwhoo/CrossAug}, a contrastive data augmentation method for debiasing fact verification models. 
Specifically, we employ a two-stage augmentation pipeline to generate new claims and evidences from existing samples.
The generated samples are then paired cross-wise with the original pair, forming contrastive samples that facilitate the model to rely less on spurious patterns and learn more robust representations.
Experimental results show that our method outperforms the previous state-of-the-art debiasing technique by 3.6\% on the debiased extension of the FEVER dataset, with a total performance boost of 10.13\% from the baseline. 
Furthermore, we evaluate our approach in data-scarce settings, where models can be more susceptible to biases due to the lack of training data. 
Experimental results demonstrate that our approach is also effective at debiasing in these low-resource conditions, exceeding the baseline performance on the Symmetric dataset with just 1\% of the original data.
\end{abstract}



\maketitle

\section{Introduction}
The ever-increasing amount of unverified information online makes it challenging to judge what to believe or distrust. 
Fact verification, the task of identifying whether a textual claim is supported or refuted by the given evidence text, can play a critical role in recognizing and correcting false information. 
Consequently, it has drawn lots of attention from the NLP community to promote the veracity and correctness of factual claims.

\begin{figure}[t]
\small
\centering
\includegraphics[width=0.85\columnwidth]{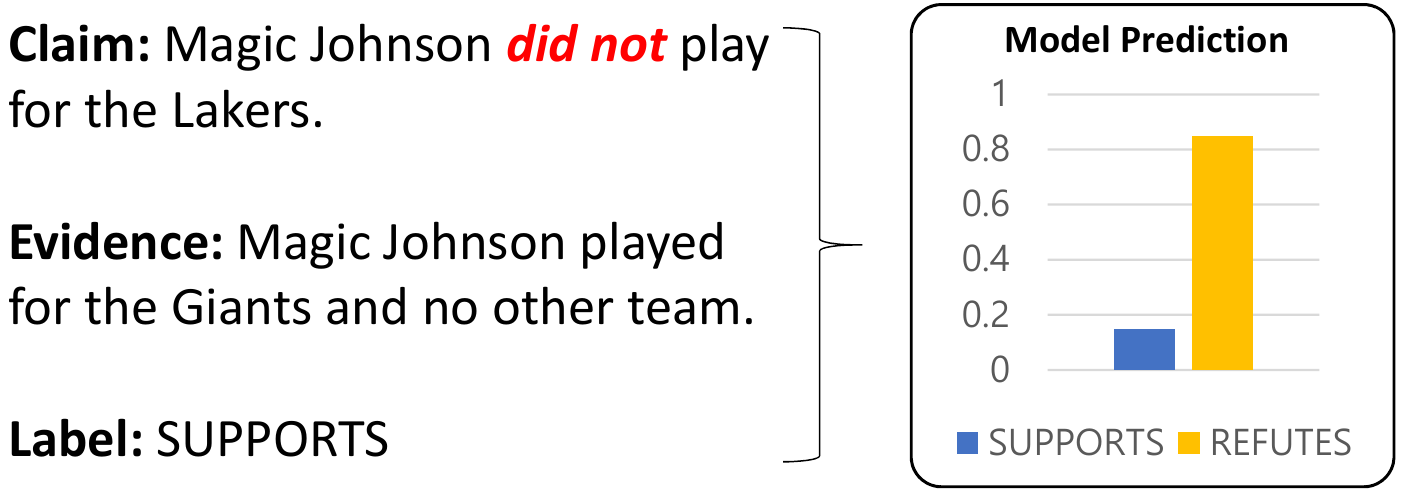}
\caption{
An illustration of incorrect prediction by a model affected by the lexical bias in FEVER dataset.
}
\vspace{-3mm}
\label{fig_prior}
\end{figure}

\begin{figure*}[t]
\small
\centering
\includegraphics[width=1.6\columnwidth]{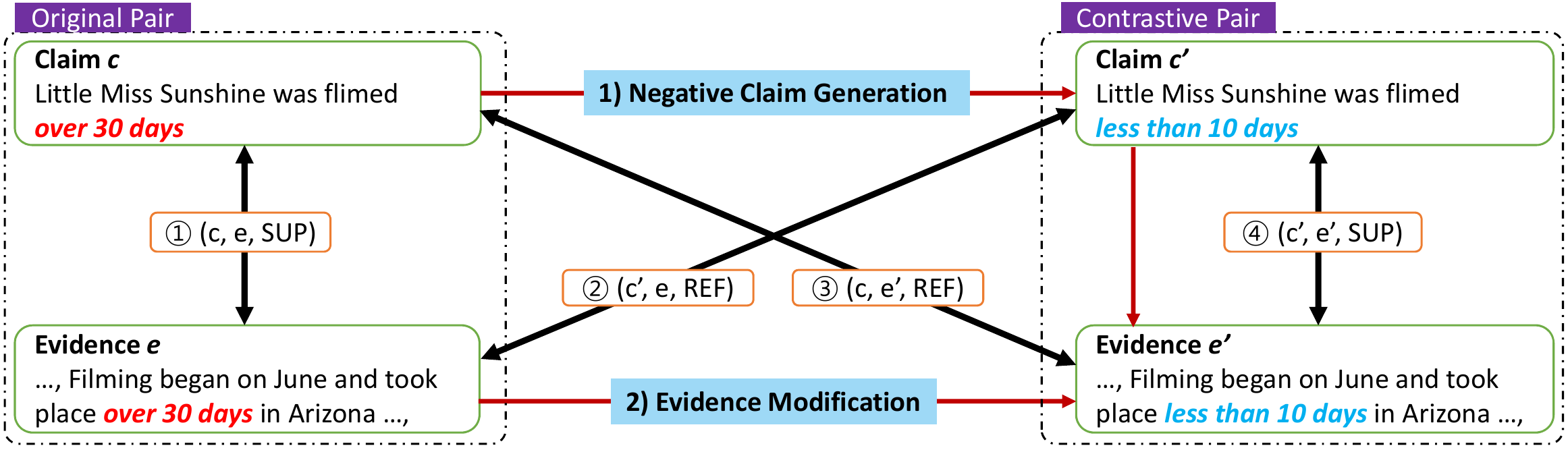}
\caption{
An overview of our data augmentation method. The pipeline consists of two stages: 1) negative claim generation and 2) evidence modification. The generated claims and evidences are paired cross-wise with the original pair to generate contrastive samples.
}
\label{fig_overall}
\vspace{-3mm}
\end{figure*}

Significant advances have been made in automated fact verification thanks to large-scale datasets \cite{Wang2017LiarLP, thorne2018fever} and pre-trained language models, such as BERT \cite{devlin2019bert}, leading to improvements in which more complex claims can be accurately fact-checked. However, recent works have demonstrated that the process of data collection using crowdsourcing often introduces idiosyncratic biases due to annotation artifacts \cite{Gururangan2018AnnotationAI, geva2019we, schuster2019towards}. These biases are typically characterized as superficial surface patterns that are strongly associated with target labels. As an example, in the FEVER dataset, negation phrases such as ``\textit{did not}'' and ``\textit{failed to}'' in the claim are highly correlated with the REFUTES label, irrespective of the given evidence \cite{schuster2019towards}.

As a result of such biases, models tend to exploit the spurious patterns between shortcut words and labels in the dataset instead of performing factual reasoning over the given evidence, as depicted in Figure~\ref{fig_prior}. In turn, models often appear to perform well on in-domain evaluation sets but show substantial performance degradation on out-of-distribution samples. Moreover, this behavior makes models vulnerable against adversarial sets, consisting of counterexamples that cause classification errors in the existing systems ~\cite{Thorne2019adversarial}.
Therefore, overcoming such biases is a key challenge in developing robust fact verification models.


To tackle this issue, previous methods either reduce the importance of the biased examples via a modified training objective \cite{mahabadi2020end}, regularize the confidence of the model on biased examples \cite{utama2020mind}, or train a model in an ensemble with a biased model to discourage it from leveraging statistical shortcuts \cite{clark2019dont}. \tony{However, the majority of these methods \minwoo{target} a specific bias; \minwoo{as a result,} they achieve improvement on the targeted set while generally resulting in poor performance on the evaluation sets that include different types of biases.
}

In this paper, we propose \textbf{CrossAug}, an alternative approach for debiasing fact verification models by augmenting the data with contrastive samples.
CrossAug generates new data samples through a novel two-stage augmentation pipeline: 1) neural-based negative claim generation and 2) lexical search-based evidence modification.
The generated claim and evidence are then paired cross-wise with the original pair, yielding contrastive pairs that are subtly different from one another in respect to context but are assigned opposite labels.
We postulate that such contrastive samples encourage a model to rely less on spurious correlations, leading to better reasoning capabilities by learning more robust representations for the task.

Indeed, our approach outperforms the regularization-based state-of-the-art method by 3.6\% on the Symmetric FEVER dataset \cite{schuster2019towards}, an unbiased evaluation set, and also shows a consistent performance boost on other fact verification datasets.


\hwanhee{To verify the performance of the proposed debiasing method on real world application scenarios where fact verification datasets often have limited amount of data, we further experiment on data-scarce settings by sub-sampling the FEVER set. Experimental results demonstrate that our approach is also effective at debiasing in these low-resource conditions, exceeding the baseline performance on the Symmetric dataset with just 1\% of the original data.}

\smallskip
In summary, our contributions in this work are as follows:
\begin{itemize}[noitemsep,leftmargin=*]
\item We propose \textbf{CrossAug}, a novel contrastive data augmentation method for debiasing fact verification models.
\item We empirically show that training a model with the data augmented by our proposed method leads to the state-of-the-art performance on the Symmetric FEVER dataset.
\item Our augmentation-based debiasing approach show performance improvements particularly in low-resource conditions compared to previous regularization-based debiasing approaches.
\end{itemize}

\section{Method}
\subsection{Task Formulation}
Given a textual claim $c$ and evidence $e$, the objective of the fact verification task is to predict whether the claim is supported by the evidence, refuted by the evidence, or the evidence has not enough information for verification. 
We denote a sample in the dataset $\mathcal{D}$ of size $N$ as a triplet $(c,e,y) \in \mathcal{D}$, where $y \in \{SUP, REF, NEI\}$ is the data label.

\begin{table*}
  \caption{Experimental results on fact verification datasets. The mean and the standard deviation of the classification accuracy over five different runs are reported for each method.}
  \vspace{-2mm}
  \label{tab:full_result}
  \begin{tabular}{l|cc|cc|cc}
    \toprule
    \textbf{Train method} & \textbf{FEVER dev} & \textbf{Symmetric} & \textbf{Adversarial} & \textbf{FM2 dev} & \textbf{$\Delta$ sym.} & \textbf{$\Delta$ avg.} \\ 
    \midrule
    No augmentation (baseline) & 86.15 $\pm$ 0.42 & 58.77 $\pm$ 1.29 & 49.66 $\pm$ 0.37 & 40.81 $\pm$ 0.43 & - & - \\
    EDA & 85.09 $\pm$ 0.25 & 58.55 $\pm$ 1.63 & 51.41  $\pm$ 1.14 & 41.21 $\pm$ 1.11 & $-$0.22\% & +0.22\% \\
    Paraphrasing & 84.33 $\pm$ 0.34 & 59.02 $\pm$ 1.38 &  \textbf{52.53} $\pm$ 1.20 & 40.60 $\pm$ 0.71 & +0.25\% & +0.27\% \\
    \midrule
    Re-weighting & 85.56 $\pm$ 0.32 & 61.87 $\pm$ 1.16 & 49.92 $\pm$ 0.80 & 43.80 $\pm$ 0.46 & +3.10\% & +1.44\% \\
    Product of Experts (PoE) & \textbf{86.50} $\pm$ 0.35 & 65.30 $\pm$ 1.73 & 51.07 $\pm$ 1.20 & \textbf{46.69 $\pm$ 1.11} & +6.53\% & +3.54\% \\
    \midrule
    CrossAug (ours) & 85.34 $\pm$ 0.68 & \textbf{68.90 $\pm$ 1.68} &51.78 $\pm$ 1.02 & 44.17 $\pm$ 1.27 & \textbf{+10.13\%} & \textbf{+3.70\%} \\
     - Negative claim only augmentation & 85.70 $\pm$ 0.28 & 61.00 $\pm$ 0.71 & 51.96 $\pm$ 0.90 & 43.06 $\pm$ 0.40 & +2.23\% & +1.58\% \\
     - Negative evidence only augmentation & 85.87 $\pm$ 0.16 & 67.06 $\pm$ 0.99 & 51.46 $\pm$ 0.43 & 43.70 $\pm$ 0.97 & +8.29\% & +3.18\% \\
    \bottomrule
  \end{tabular}
  \vspace{-1mm}
\end{table*}

\subsection{Data augmentation pipeline}
In our proposed method, we generate three additional synthetic samples for each original claim-evidence pair through two stages of augmentation. Note that CrossAug utilizes only positive claims (SUP claims), which are verifiable by specific evidence, in the FEVER dataset. The whole process of our data augmentation pipeline is shown in Figure \ref{fig_overall}.

\noindent\textbf{(1) Negative Claim Generation:}
The first stage is to generate a \textit{negative claim} $c'$ based on a positive claim $c$ by adopting a neural sequence-to-sequence model. This generative process involves transforming of a positive claim, such as inserting a negation or replacing a word with antonyms. 
\tony{In turn, the generated claim has a different meaning so that it is refuted by the evidence $e$ supporting a positive claim, which is why we call this negative claim.}
Through this process, we form a new data sample $(c', e, REF)$.

To this end, we fine-tune BART \cite{lewis2020bart} on WikiFactCheck-English dataset, which provides pairs of positive claims and their corresponding negative claims \cite{sathe2020automated}. In fine-tuning, positive claims are used as the source text and negative claims are taken as the target text of the model. To provide a richer context, we also fine-tune the model with the source and target text reversed since a positive claim can be seen as a refuted version of a negative claim.


\noindent\textbf{(2) Evidence Modification:}
The negative claim generated in the first stage often only differs from the positive claim by a few words, and can thus be seen as a span replacement. For example, ``\textit{over 30 days}" is simply substituted with ``\textit{less than 10 days}", as shown in Figure \ref{fig_overall}. This phenomenon is due to the characteristics of the data used for fine-tuning the model: the negative claim was manually written by annotators under the constraints of both the sentence length and the subject to be similar to the positive claim \cite{sathe2020automated}. We also observe that the words replaced in the positive claim are often found verbatim in the evidence. This is because the changed part of the claim usually matches the factual information taken from the evidence.

As the second stage of our data augmentation process, we build upon these observations to perform a lexical search-based evidence modification. First, we compare the positive claim $c$ with the negative claim $c'$ to identify the changed part in the first stage. Once the words replaced in the positive claim are recognized (``over 30 days" in Figure \ref{fig_overall}), we search for the same words in the evidence $e$ and replace them with the substituted words in the negative claim (``less than 10 days" in Figure \ref{fig_overall}). Since this substitution induces the same factual modification on the evidence as that applied to the negative claim, it can be logically concluded that the resulting modified evidence $e'$ supports the negative claim $c'$ and refutes the positive claim $c$. 
Consequently, we form two additional contrastive samples $(c, e', REF)$ and $(c', e', SUP)$ in the second stage.


\smallskip
\noindent\textbf{Exceptional Cases:}
In the first stage, the generated negative claim $c'$ is occasionally the exact same as the positive claim $c$. For such samples, we skip performing augmentation.
Also, in the second stage, we carry out the evidence modification only when the number of replaced words in the claims are less than or equal to $\tau$, where $\tau$ is a threshold value. This is necessary to prevent invalid evidence modification. When the replaced part is large, it frequently contains inappropriate terms for reconstructing the evidence, such as non-factual words, producing an illogical sentence. However, we still keep the sample $(c',e, REF)$ from the first stage even when the evidence modification stage is skipped.

\section{Experiments}

For our experiments, we evaluate our proposed data augmentation method on four datasets, including FEVER, and compare its performance with the existing methods. 


\subsection{Datasets}
\textbf{FEVER} \cite{thorne2018fever} is a crowdsourced fact verification dataset containing claim-evidence pairs based on Wikipedia articles. We only use the claims paired with a single evidence for training and evaluation in this work.

\noindent\textbf{Symmetric} \cite{schuster2019towards} is a test set based on the FEVER development dataset designed for unbiased evaluation. It eliminates the correlation of n-grams in the claim and labels by careful construction.


\noindent\textbf{Adversarial} \cite{Thorne2019adversarial} is an adversarially constructed dataset explicitly designed to induce errors in models trained on the FEVER dataset.


\noindent\textbf{Fool Me Twice (FM2)} \cite{eisenschlos2021fool} is a Wikipedia-based fact verification dataset composed of 13k claim-evidence pairs that are collected through the games among crowd-workers. 

\subsection{Compared Methods}

\textbf{Data Augmentation Methods:}  
We compare our method against two data augmentation techniques commonly used for various natural language processing tasks: Easy Data Augmentation (EDA) \cite{wei2019eda} and neural paraphrasing. EDA applies simple mutations, such as random swapping or synonym replacement, to the original sentence to generate new examples. 
For neural paraphrasing, we use a GPT-2 model \cite{radford2019language} fine-tuned on back-translated data to paraphrase the original text \cite{krishna-etal-2020-reformulating}. For each original claim-evidence pair, we create a new pair that holds the same relation by transforming only the claim using these methods, leading to the augmentation ratio of augmented to original data 1:1.

\noindent\textbf{Regularization-based Debiasing Methods:}
We also compare with
two debiasing techniques that reduce the reliance on biases by regularizing the model on the biased samples.
The first one is an example re-weighting method that targets biases from the shortcut words \cite{schuster2019towards}. By re-weighting the importance of claims containing those words, it forces a model to focus on the hard examples in which relying on the bias results in incorrect predictions.
The other one is Product of Experts (PoE) \cite{mahabadi2020end}, which computes the training loss in an ensemble of the base model and the bias-only model. Similar to the first method, it controls the base model's loss depending on the prediction of the bias-only model for each example.



\subsection{Implementation Details}
For our experiments, we use the BERT-base-uncased model \cite{devlin2019bert}, which demonstrates competitive performance for fact verification tasks. 
We fine-tune BERT with an additional layer on top of the [CLS] token embedding.
We concatenate the claim and the evidence, and insert [SEP] token in between to make the input sequence.
Following previous works, we set maximum sequence length to 128, batch size to 32, and optimize the model through a standard cross-entropy loss using the Adam optimizer~\cite{kingma2014adam} with a learning rate of 2e-5.
\minwoo{We train the model on the FEVER train set and evaluate the generalization performance using the development set for Symmetric, Adversarial and FM2 datasets.}
We train the model for 3 epochs with 5 different random seeds for all experiments and report the averaged result.
For our augmentation pipeline, we set the maximum span size for evidence modification $\tau$ to $3$, which produces an augmented dataset with an augmentation ratio of 1:0.58.

\subsection{Results on the Full Dataset}

First we augment the full FEVER train set with our approach and compare its performance in Table~\ref{tab:full_result}. 
Our proposed method achieves 10.13\% improvement over the baseline and 3.6\% improvement over the previous state-of-the-art debiasing technique on the Symmetric dataset. 
This result shows that our method is highly effective at preventing the model to predict from unnecessary biases.
Our approach also shows a 2.12\% improvement on the Adversarial dataset and 3.36\% improvement on the FM2 dataset compared to the baseline, indicating that our augmentation method benefits not just on the diagnostic dataset for lexical bias but for fact verification in general.
Finally, our method leads to the greatest overall improvement across the datasets out of all compared training methods,
This result empirically proves that the contrastive samples generated from our augmentation method enhance the factual reasoning capabilities by learning a more robust feature representation, achieving strong generalization.

Compared to our approach, the example re-weighting and PoE methods perform slightly worse on the Symmetric and Adversarial dataset and slightly better on the original FEVER development set and FM2 dataset. 
On the other hand, EDA and paraphrasing augmentations show negligible performance improvement on the Symmetric dataset. These results suggest that simply training with more data does not necessarily help mitigate the bias in data. 

\subsection{Ablation Studies}

We conduct ablation study to verify the effectiveness of the augmented samples generated from each step of our augmentation process.
The results shown in Table \ref{tab:full_result} reveal that a moderate performance improvement on the Symmetric, Adversarial and FM2 evaluation sets are attained even with only using the negative claims generated from the first step. 
However, its performance on the Symmetric dataset is still significantly lower compared to the full augmentation method, implying that augmenting with negative claims alone is less effective for debiasing the model. 

Training with the negative evidence augmented data exhibits a more competitive performance on the other hand, especially outperforming previous state-of-the-art technique on the Symmetric dataset. 
The results imply that the key component of our debiasing approach is training the model with contrastive samples sharing the same claim, which enables the model to learn by comparing the claim to the input evidence instead of just relying on artifacts in the claim.
Nevertheless, the full augmentation method outperforms all ablations, indicating that contrastive data samples in general help the model learn more robust representations.

\subsection{Results on Low-resource Conditions}\label{SEC:LOW}

\minwoo{In the real world, the available training corpus size for fact verification is often lacking due to the expensive cost and difficulty of collecting and labelling the data.
Training on such a limited amount of data could lead to even more biased models due to the lack of samples to learn factual reasoning from.
Therefore, it is important to investigate whether the debiasing methods perform well in low-resource conditions.
}

\begin{figure}[t]
    \begin{center}
    \begin{tabular}{c}
      \includegraphics[width=0.8\columnwidth]{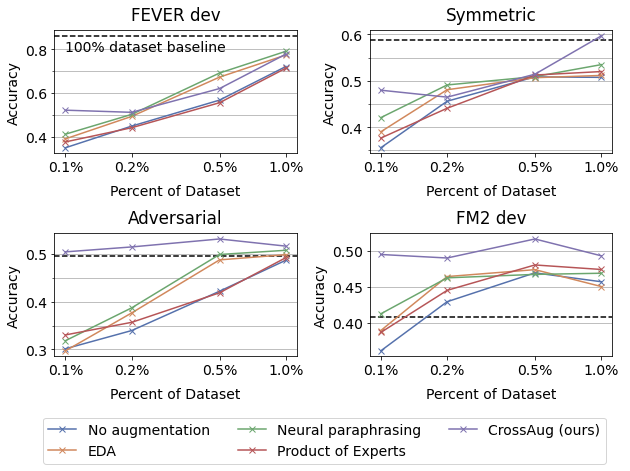}
    \end{tabular}
    \vspace{-2mm}
    \caption{Experimental results on varying training set size with different training methods. Train set sampling is done with five different seeds and the mean accuracy is reported. 
    }
    \vspace{-4mm}
    \label{fig:limited}
    \end{center}
\end{figure}

Thus, we further evaluate our data augmentation method in low-resource conditions, simulating a data-scarce setting by sampling subsets of the original FEVER training data. 
We perform a class-balanced sub-sample on the FEVER training data with 5 different random seeds, and train the model on each subset with another 5 different random seeds to account for statistical variations.
As in the previous experiments, we augment the subset with an augmentation ratio of 1:1 for EDA and back-translation methods, while our augmentation method results in an average augmentation ratio of 1:0.47.
We evaluate on the subsets of 0.1\% ($N=240$), 0.2\% ($N=483$), 0.5\% ($N=969$), and 1.0\% ($N=2,427$) of the original training data, and the results are presented in Figure \ref{fig:limited}. \\
\indent Our augmentation method shows a consistent improvement in the low-resource conditions over all evaluated datasets, and especially outperforms the baseline trained on the full dataset with just 1\% of the original training data for the Symmetric evaluation set. On the other hand, the PoE method shows little to no improvements on all of the datasets except FM2, and in some cases show a slight performance drop. 
These results indicate that PoE, which relies on training a biased model to regularize learning from biased samples, do not generalize well in data-scarce settings.\\
EDA and paraphrasing augmentation show a moderate improvement over the baselines, verifying their effectiveness in low-resource conditions. 
However, our augmentation method is still able to achieve a marked improvement over EDA and paraphrasing on Adversarial and FM2 datasets, implying that our augmentation approach is more robust and effective at generalizing to out-of-distribution samples. 
In summary, the results with varying training dataset sizes show that our augmentation method is effective for low-resource domains.

\section{Conclusion}
In this work, we propose a novel data augmentation method for debiasing fact verification models.
Our approach consists of generating negative claim and evidence pairs and forming contrastive samples to augment the data,
which facilitates the training model to rely less on the spurious correlations and learn better representations.
We evaluate our approach on various fact verification datasets and show that our method outperforms previous methods on the unbiased evaluation set. 
We also show that our approach is effective in low-resource conditions with limited data compared to regularization-based debiasing approaches.

\begin{acks}
Kyomin Jung is with ASRI, Seoul National University, Seoul, Korea. This work was supported by AIRS Company in Hyundai Motor and Kia through HMC/KIA-SNU AI Consortium Fund.
\end{acks}

\clearpage
\bibliographystyle{ACM-Reference-Format}
\bibliography{biblio}


\begin{thebibliography}{17}


\ifx \showCODEN    \undefined \def \showCODEN     #1{\unskip}     \fi
\ifx \showDOI      \undefined \def \showDOI       #1{#1}\fi
\ifx \showISBNx    \undefined \def \showISBNx     #1{\unskip}     \fi
\ifx \showISBNxiii \undefined \def \showISBNxiii  #1{\unskip}     \fi
\ifx \showISSN     \undefined \def \showISSN      #1{\unskip}     \fi
\ifx \showLCCN     \undefined \def \showLCCN      #1{\unskip}     \fi
\ifx \shownote     \undefined \def \shownote      #1{#1}          \fi
\ifx \showarticletitle \undefined \def \showarticletitle #1{#1}   \fi
\ifx \showURL      \undefined \def \showURL       {\relax}        \fi
\providecommand\bibfield[2]{#2}
\providecommand\bibinfo[2]{#2}
\providecommand\natexlab[1]{#1}
\providecommand\showeprint[2][]{arXiv:#2}

\bibitem[\protect\citeauthoryear{Clark, Yatskar, and Zettlemoyer}{Clark
  et~al\mbox{.}}{2019}]%
        {clark2019dont}
\bibfield{author}{\bibinfo{person}{Christopher Clark}, \bibinfo{person}{Mark
  Yatskar}, {and} \bibinfo{person}{Luke Zettlemoyer}.}
  \bibinfo{year}{2019}\natexlab{}.
\newblock \showarticletitle{Don{'}t Take the Easy Way Out: Ensemble Based
  Methods for Avoiding Known Dataset Biases}. In
  \bibinfo{booktitle}{\emph{Proceedings of the 2019 Conference on Empirical
  Methods in Natural Language Processing and the 9th International Joint
  Conference on Natural Language Processing (EMNLP-IJCNLP)}}.
  \bibinfo{publisher}{Association for Computational Linguistics},
  \bibinfo{address}{Hong Kong, China}, \bibinfo{pages}{4069--4082}.
\newblock
\urldef\tempurl%
\url{https://doi.org/10.18653/v1/D19-1418}
\showDOI{\tempurl}


\bibitem[\protect\citeauthoryear{Devlin, Chang, Lee, and Toutanova}{Devlin
  et~al\mbox{.}}{2019}]%
        {devlin2019bert}
\bibfield{author}{\bibinfo{person}{Jacob Devlin}, \bibinfo{person}{Ming-Wei
  Chang}, \bibinfo{person}{Kenton Lee}, {and} \bibinfo{person}{Kristina
  Toutanova}.} \bibinfo{year}{2019}\natexlab{}.
\newblock \showarticletitle{{BERT}: Pre-training of Deep Bidirectional
  Transformers for Language Understanding}. In
  \bibinfo{booktitle}{\emph{Proceedings of the 2019 Conference of the North
  {A}merican Chapter of the Association for Computational Linguistics: Human
  Language Technologies, Volume 1 (Long and Short Papers)}}.
  \bibinfo{publisher}{Association for Computational Linguistics},
  \bibinfo{address}{Minneapolis, Minnesota}, \bibinfo{pages}{4171--4186}.
\newblock
\urldef\tempurl%
\url{https://doi.org/10.18653/v1/N19-1423}
\showDOI{\tempurl}


\bibitem[\protect\citeauthoryear{Eisenschlos, Dhingra, Bulian, B{\"o}rschinger,
  and Boyd-Graber}{Eisenschlos et~al\mbox{.}}{2021}]%
        {eisenschlos2021fool}
\bibfield{author}{\bibinfo{person}{Julian Eisenschlos}, \bibinfo{person}{Bhuwan
  Dhingra}, \bibinfo{person}{Jannis Bulian}, \bibinfo{person}{Benjamin
  B{\"o}rschinger}, {and} \bibinfo{person}{Jordan Boyd-Graber}.}
  \bibinfo{year}{2021}\natexlab{}.
\newblock \showarticletitle{Fool Me Twice: Entailment from Wikipedia
  Gamification}. In \bibinfo{booktitle}{\emph{Proceedings of the 2021
  Conference of the North American Chapter of the Association for Computational
  Linguistics: Human Language Technologies}}. \bibinfo{pages}{352--365}.
\newblock


\bibitem[\protect\citeauthoryear{Geva, Goldberg, and Berant}{Geva
  et~al\mbox{.}}{2019}]%
        {geva2019we}
\bibfield{author}{\bibinfo{person}{Mor Geva}, \bibinfo{person}{Yoav Goldberg},
  {and} \bibinfo{person}{Jonathan Berant}.} \bibinfo{year}{2019}\natexlab{}.
\newblock \showarticletitle{Are We Modeling the Task or the Annotator? An
  Investigation of Annotator Bias in Natural Language Understanding Datasets}.
  In \bibinfo{booktitle}{\emph{Proceedings of the 2019 Conference on Empirical
  Methods in Natural Language Processing and the 9th International Joint
  Conference on Natural Language Processing (EMNLP-IJCNLP)}}.
  \bibinfo{pages}{1161--1166}.
\newblock


\bibitem[\protect\citeauthoryear{Gururangan, Swayamdipta, Levy, Schwartz,
  Bowman, and Smith}{Gururangan et~al\mbox{.}}{2018}]%
        {Gururangan2018AnnotationAI}
\bibfield{author}{\bibinfo{person}{Suchin Gururangan}, \bibinfo{person}{Swabha
  Swayamdipta}, \bibinfo{person}{Omer Levy}, \bibinfo{person}{Roy Schwartz},
  \bibinfo{person}{Samuel~R. Bowman}, {and} \bibinfo{person}{Noah~A. Smith}.}
  \bibinfo{year}{2018}\natexlab{}.
\newblock \showarticletitle{Annotation Artifacts in Natural Language Inference
  Data}. In \bibinfo{booktitle}{\emph{NAACL-HLT}}.
\newblock


\bibitem[\protect\citeauthoryear{Karimi~Mahabadi, Belinkov, and
  Henderson}{Karimi~Mahabadi et~al\mbox{.}}{2020}]%
        {mahabadi2020end}
\bibfield{author}{\bibinfo{person}{Rabeeh Karimi~Mahabadi},
  \bibinfo{person}{Yonatan Belinkov}, {and} \bibinfo{person}{James Henderson}.}
  \bibinfo{year}{2020}\natexlab{}.
\newblock \showarticletitle{End-to-End Bias Mitigation by Modelling Biases in
  Corpora}. In \bibinfo{booktitle}{\emph{Proceedings of the 58th Annual Meeting
  of the Association for Computational Linguistics}}.
  \bibinfo{publisher}{Association for Computational Linguistics},
  \bibinfo{address}{Online}, \bibinfo{pages}{8706--8716}.
\newblock
\urldef\tempurl%
\url{https://doi.org/10.18653/v1/2020.acl-main.769}
\showDOI{\tempurl}


\bibitem[\protect\citeauthoryear{Kingma and Ba}{Kingma and Ba}{2015}]%
        {kingma2014adam}
\bibfield{author}{\bibinfo{person}{Diederik~P. Kingma} {and}
  \bibinfo{person}{Jimmy Ba}.} \bibinfo{year}{2015}\natexlab{}.
\newblock \showarticletitle{Adam: {A} Method for Stochastic Optimization}. In
  \bibinfo{booktitle}{\emph{3rd International Conference on Learning
  Representations, {ICLR} 2015, San Diego, CA, USA, May 7-9, 2015, Conference
  Track Proceedings}}, \bibfield{editor}{\bibinfo{person}{Yoshua Bengio} {and}
  \bibinfo{person}{Yann LeCun}} (Eds.).
\newblock
\urldef\tempurl%
\url{http://arxiv.org/abs/1412.6980}
\showURL{%
\tempurl}


\bibitem[\protect\citeauthoryear{Krishna, Wieting, and Iyyer}{Krishna
  et~al\mbox{.}}{2020}]%
        {krishna-etal-2020-reformulating}
\bibfield{author}{\bibinfo{person}{Kalpesh Krishna}, \bibinfo{person}{John
  Wieting}, {and} \bibinfo{person}{Mohit Iyyer}.}
  \bibinfo{year}{2020}\natexlab{}.
\newblock \showarticletitle{Reformulating Unsupervised Style Transfer as
  Paraphrase Generation}. In \bibinfo{booktitle}{\emph{Proceedings of the 2020
  Conference on Empirical Methods in Natural Language Processing (EMNLP)}}.
  \bibinfo{publisher}{Association for Computational Linguistics},
  \bibinfo{address}{Online}, \bibinfo{pages}{737--762}.
\newblock
\urldef\tempurl%
\url{https://doi.org/10.18653/v1/2020.emnlp-main.55}
\showDOI{\tempurl}


\bibitem[\protect\citeauthoryear{Lewis, Liu, Goyal, Ghazvininejad, Mohamed,
  Levy, Stoyanov, and Zettlemoyer}{Lewis et~al\mbox{.}}{2020}]%
        {lewis2020bart}
\bibfield{author}{\bibinfo{person}{Mike Lewis}, \bibinfo{person}{Yinhan Liu},
  \bibinfo{person}{Naman Goyal}, \bibinfo{person}{Marjan Ghazvininejad},
  \bibinfo{person}{Abdelrahman Mohamed}, \bibinfo{person}{Omer Levy},
  \bibinfo{person}{Veselin Stoyanov}, {and} \bibinfo{person}{Luke
  Zettlemoyer}.} \bibinfo{year}{2020}\natexlab{}.
\newblock \showarticletitle{BART: Denoising Sequence-to-Sequence Pre-training
  for Natural Language Generation, Translation, and Comprehension}. In
  \bibinfo{booktitle}{\emph{Proceedings of the 58th Annual Meeting of the
  Association for Computational Linguistics}}. \bibinfo{pages}{7871--7880}.
\newblock


\bibitem[\protect\citeauthoryear{Radford, Wu, Child, Luan, Amodei, and
  Sutskever}{Radford et~al\mbox{.}}{2019}]%
        {radford2019language}
\bibfield{author}{\bibinfo{person}{Alec Radford}, \bibinfo{person}{Jeffrey Wu},
  \bibinfo{person}{Rewon Child}, \bibinfo{person}{David Luan},
  \bibinfo{person}{Dario Amodei}, {and} \bibinfo{person}{Ilya Sutskever}.}
  \bibinfo{year}{2019}\natexlab{}.
\newblock \showarticletitle{Language models are unsupervised multitask
  learners}.
\newblock \bibinfo{journal}{\emph{OpenAI blog}} \bibinfo{volume}{1},
  \bibinfo{number}{8} (\bibinfo{year}{2019}), \bibinfo{pages}{9}.
\newblock


\bibitem[\protect\citeauthoryear{Sathe, Ather, Le, Perry, and Park}{Sathe
  et~al\mbox{.}}{2020}]%
        {sathe2020automated}
\bibfield{author}{\bibinfo{person}{Aalok Sathe}, \bibinfo{person}{Salar Ather},
  \bibinfo{person}{Tuan~Manh Le}, \bibinfo{person}{Nathan Perry}, {and}
  \bibinfo{person}{Joonsuk Park}.} \bibinfo{year}{2020}\natexlab{}.
\newblock \showarticletitle{Automated Fact-Checking of Claims from Wikipedia}.
  In \bibinfo{booktitle}{\emph{Proceedings of The 12th Language Resources and
  Evaluation Conference}}. \bibinfo{pages}{6874--6882}.
\newblock


\bibitem[\protect\citeauthoryear{Schuster, Shah, Yeo, Roberto Filizzola~Ortiz,
  Santus, and Barzilay}{Schuster et~al\mbox{.}}{2019}]%
        {schuster2019towards}
\bibfield{author}{\bibinfo{person}{Tal Schuster}, \bibinfo{person}{Darsh Shah},
  \bibinfo{person}{Yun Jie~Serene Yeo}, \bibinfo{person}{Daniel Roberto
  Filizzola~Ortiz}, \bibinfo{person}{Enrico Santus}, {and}
  \bibinfo{person}{Regina Barzilay}.} \bibinfo{year}{2019}\natexlab{}.
\newblock \showarticletitle{Towards Debiasing Fact Verification Models}. In
  \bibinfo{booktitle}{\emph{Proceedings of the 2019 Conference on Empirical
  Methods in Natural Language Processing and the 9th International Joint
  Conference on Natural Language Processing (EMNLP-IJCNLP)}}.
  \bibinfo{publisher}{Association for Computational Linguistics},
  \bibinfo{address}{Hong Kong, China}, \bibinfo{pages}{3419--3425}.
\newblock
\urldef\tempurl%
\url{https://doi.org/10.18653/v1/D19-1341}
\showDOI{\tempurl}


\bibitem[\protect\citeauthoryear{Thorne and Vlachos}{Thorne and
  Vlachos}{2019}]%
        {Thorne2019adversarial}
\bibfield{author}{\bibinfo{person}{James Thorne} {and} \bibinfo{person}{Andreas
  Vlachos}.} \bibinfo{year}{2019}\natexlab{}.
\newblock \bibinfo{title}{Adversarial attacks against {Fact Extraction and
  VERification}}.
\newblock
\newblock
\showeprint[arxiv]{1903.05543}~[cs.CL]


\bibitem[\protect\citeauthoryear{Thorne, Vlachos, Christodoulopoulos, and
  Mittal}{Thorne et~al\mbox{.}}{2018}]%
        {thorne2018fever}
\bibfield{author}{\bibinfo{person}{James Thorne}, \bibinfo{person}{Andreas
  Vlachos}, \bibinfo{person}{Christos Christodoulopoulos}, {and}
  \bibinfo{person}{Arpit Mittal}.} \bibinfo{year}{2018}\natexlab{}.
\newblock \showarticletitle{{FEVER}: a Large-scale Dataset for Fact Extraction
  and {VER}ification}. In \bibinfo{booktitle}{\emph{Proceedings of the 2018
  Conference of the North {A}merican Chapter of the Association for
  Computational Linguistics: Human Language Technologies, Volume 1 (Long
  Papers)}}. \bibinfo{publisher}{Association for Computational Linguistics},
  \bibinfo{address}{New Orleans, Louisiana}, \bibinfo{pages}{809--819}.
\newblock
\urldef\tempurl%
\url{https://doi.org/10.18653/v1/N18-1074}
\showDOI{\tempurl}


\bibitem[\protect\citeauthoryear{Utama, Moosavi, and Gurevych}{Utama
  et~al\mbox{.}}{2020}]%
        {utama2020mind}
\bibfield{author}{\bibinfo{person}{Prasetya~Ajie Utama},
  \bibinfo{person}{Nafise~Sadat Moosavi}, {and} \bibinfo{person}{Iryna
  Gurevych}.} \bibinfo{year}{2020}\natexlab{}.
\newblock \showarticletitle{Mind the Trade-off: Debiasing NLU Models without
  Degrading the In-distribution Performance}. In
  \bibinfo{booktitle}{\emph{Proceedings of the 58th Annual Meeting of the
  Association for Computational Linguistics}}. \bibinfo{pages}{8717--8729}.
\newblock


\bibitem[\protect\citeauthoryear{Wang}{Wang}{2017}]%
        {Wang2017LiarLP}
\bibfield{author}{\bibinfo{person}{William~Yang Wang}.}
  \bibinfo{year}{2017}\natexlab{}.
\newblock \showarticletitle{"Liar, Liar Pants on Fire": A New Benchmark Dataset
  for Fake News Detection}. In \bibinfo{booktitle}{\emph{ACL}}.
\newblock


\bibitem[\protect\citeauthoryear{Wei and Zou}{Wei and Zou}{2019}]%
        {wei2019eda}
\bibfield{author}{\bibinfo{person}{Jason Wei} {and} \bibinfo{person}{Kai Zou}.}
  \bibinfo{year}{2019}\natexlab{}.
\newblock \showarticletitle{EDA: Easy Data Augmentation Techniques for Boosting
  Performance on Text Classification Tasks}. In
  \bibinfo{booktitle}{\emph{Proceedings of the 2019 Conference on Empirical
  Methods in Natural Language Processing and the 9th International Joint
  Conference on Natural Language Processing (EMNLP-IJCNLP)}}.
  \bibinfo{pages}{6383--6389}.
\newblock


\end{thebibliography}

\end{document}